\documentclass[10pt,twocolumn,letterpaper]{article}

\usepackage{cvpr}              
\usepackage{algorithm}
\usepackage{algorithmicx}
\usepackage{algpseudocode}
\definecolor{cvprblue}{rgb}{0.21,0.49,0.74}
\usepackage[pagebackref,breaklinks,colorlinks,allcolors=cvprblue]{hyperref}

\def\paperID{*****} 
\def\confName{CVPR}
\def\confYear{2026}

\title{Med-CAM: Minimal Evidence for Explaining Medical Decision Making}

\author{Pirzada Suhail\\
IIT Bombay\\
{\tt\small psuhail@iitb.ac.in}
\and
Aditya Anand\\
IIT Bombay\\
{\tt\small adityaanand@iitb.ac.in}
\and
Amit Sethi\\
IIT Bombay\\
{\tt\small asethi@iitb.ac.in}
}

\begin{document}
\maketitle
\begin{abstract}

Reliable and interpretable decision-making is essential in medical imaging, where diagnostic outcomes directly influence patient care. Despite advances in deep learning, most medical AI systems operate as opaque black boxes, providing little insight into \emph{why} a particular diagnosis was reached. In this paper, we introduce \textbf{Med-CAM}, a framework for generating minimal and sharp maps as evidence-based explanations for \textbf{Med}ical decision making via \textbf{C}lassifier \textbf{A}ctivation \textbf{M}atching. \textbf{Med-CAM} trains a segmentation network from scratch to produce a mask that highlights the minimal evidence critical to model’s decision for any seen or unseen image. This ensures that the explanation is both \emph{faithful} to the network’s behavior and \emph{interpretable} to clinicians. Experiments show, unlike prior spatial explanation methods, such as Grad-CAM and attention maps, which yield only fuzzy regions of relative importance, Med-CAM with its superior spatial awareness to shapes, textures, and boundaries, delivers conclusive, evidence-based explanations that faithfully replicate the model’s prediction for any given image. By explicitly constraining explanations to be compact, consistent with model activations, and diagnostic alignment, MedCAM advances transparent AI to foster clinician understanding and trust in high-stakes medical applications such as pathology and radiology.

\end{abstract}

\section{Introduction}
\label{sec:intro}

Explanations are increasingly being recognized as essential for understanding and trusting the decision-making of machine learning systems deployed in medical settings. In clinical workflows, where algorithmic predictions can directly influence diagnosis or treatment, it is not enough for a model to be accurate—it must also be interpretable and verifiable. Deep neural networks, despite their remarkable success across medical imaging tasks such as pathology, radiology, and dermatology, often operate as opaque black boxes. Their predictions arise from complex, high-dimensional computations that are inaccessible to human reasoning, leaving clinicians uncertain about whether the model’s conclusions stem from meaningful pathology or from confounding artifacts.

In this context, \emph{minimal evidence} becomes a key desideratum for trustworthy medical explanations. By isolating the smallest set of image regions sufficient to support a model’s diagnostic decision, one can reveal the true basis of the model’s reasoning in a form that is both human-understandable and clinically verifiable. Minimal explanations reveal the essential visual cues such as atypical cellular morphologies in histopathology or subtle opacities in radiographs, that directly drive the network’s prediction. Such evidence-centered explanations not only enhance clinical interpretability but also serve as a diagnostic sanity check, allowing practitioners to verify weather that the model attends to medically relevant structures or not.

To address this need, we propose \textbf{Med-CAM}, a \textbf{Med}ical \textbf{C}lassifier \textbf{A}ctivation \textbf{M}atching framework for generating \textbf{minimal, evidence-based explanations} for medicinal decision making. Med-CAM learns a lightweight U-Net from scratch that produces a binary mask highlighting the minimal evidence necessary for a model’s output. The U-Net is trained individually on a single image within seconds, optimizing for consistency between the classifier’s activations on the original image and on the masked explanation. This \emph{activation-matching} principle ensures that the explanation preserves both the internal reasoning and the diagnostic prediction of the underlying model.

Compared to existing saliency-based approaches such as Grad-CAM and attention maps, Med-CAM offers several key advantages. Grad-CAM provides gradient-weighted heatmaps that highlight regions of relative importance but fail to capture precise spatial characteristics such as boundaries, textures, or surface continuity. In contrast, Med-CAM explicitly enforces activation and decision consistency, resulting in spatially aware, compact, and conclusive evidence maps that faithfully replicate the model’s diagnostic behavior. By grounding explanations in minimal, activation-consistent evidence, Med-CAM bridges the gap between model transparency and clinical interpretability, contributing toward the development of trustworthy medical AI.

\section{Prior Work}

\subsection{Inversion}
Inversion methods aim to reconstruct inputs that elicit specific outputs or internal activations of a neural network seeking to visualize what a model has learned rather than why it makes a particular decision. Early work reconstructed representative patterns from multilayer perceptrons through gradient-based optimization, though such visualizations were often noisy or adversarial-like~\cite{KINDERMANN1990277,784232,SAAD200778}. Subsequent studies explored evolutionary search and constrained optimization~\cite{Wong2017NeuralNI}, followed by the introduction of prior-based regularization techniques such as smoothness constraints and pretrained generative models to enhance realism and interpretability~\cite{mahendran2015understanding,yosinski2015understanding,mordvintsev2015inceptionism,nguyen2016synthesizing,nguyen2017plug}. More recent methods stabilize the inversion process through learned surrogate landscapes~\cite{liu2022landscapelearningneuralnetwork} or reframe it within logical reasoning frameworks for deterministic reconstruction~\cite{suhail2024network} while others \cite{suhail2024networkcnn} use generative modeling for inversion.

\subsection{Explainability}
While inversion captures model behavior in aggregate, explainability focuses on generating faithful rationales for individual predictions. Explainable AI (XAI) has therefore emerged as a major research area~\cite{ALI2023101805,hsieh2024comprehensiveguideexplainableai,Gilpin2018ExplainingEA}, driven by the need to enhance trust, transparency, and accountability in high-stakes applications.  

Post-hoc attribution methods remain the dominant paradigm. \textbf{LIME} builds local surrogate models to approximate black-box predictors~\cite{10022096}, while \textbf{Anchors}~\cite{10.5555/3504035.3504222} extends this idea through high-precision, rule-based local explanations that define sufficient conditions for predictions. \textbf{Grad-CAM} highlights salient image regions by gradient-weighted class activation mapping~\cite{Selvaraju_2019}, whereas \textbf{DeepLIFT}~\cite{shrikumar2019learningimportantfeaturespropagating} backpropagates contribution scores relative to reference activations. The \textbf{SHAP} framework~\cite{lundberg2017unifiedapproachinterpretingmodel} unifies several additive feature-attribution approaches using Shapley values to assign consistent feature importance scores.  

Beyond local attributions, concept-based and structural methods attempt to align neural activations with semantically meaningful parts~\cite{10.1007/978-3-031-92648-8_17}. Architectural approaches ~\cite{böhle2022bcosnetworksalignmentneed} promote explicit weight–input alignment for interpretable transformations, while ~\cite{stalder2022classifyblackboxattributions} introduces an auxiliary explainer network that produces class-specific binary masks identifying discriminative evidence regions.  

Parallel to these efforts, logic- and reasoning-based frameworks explore explanation as a formal process. ~\cite{darwiche2020reasonsdecisions} introduced theoretical foundations for necessary, sufficient, and complete reasons behind classifier decisions. Similarly, ~\cite{10.5555/3304652.3304719} proposed a symbolic approach to explain Bayesian network classifiers, identifying minimal sufficient feature sets responsible for classification. Abductive reasoning techniques~\cite{ignatiev2018abductionbasedexplanationsmachinelearning} extend this logic-driven line of work by computing subset- or cardinality-minimal explanations with formal guarantees of faithfulness. Recent surveys~\cite{electronics10050593} emphasize the need for quantitative evaluation criteria—combining fidelity, stability, and human-centered assessment—to move beyond purely visual interpretability. Interactive frameworks like ~\cite{teso2022leveragingexplanationsinteractivemachine} further demonstrate how explanations can guide human-in-the-loop model refinement. 

\subsection{Explainability in Medical Diagnostics}
In medical imaging, interpretability is not merely desirable—it is essential. ~\cite{singh2020explainabledeeplearningmodels} surveyed explainable deep learning methods for medical image analysis and highlighted major barriers to clinical adoption, including lack of interpretive rigor and standardized evaluation. ~\cite{VANDERVELDEN2022102470} proposed a comprehensive taxonomy of explainable medical imaging methods categorized by anatomical region, modality, and transparency level. More recently, ~\cite{Chaddad_2025} emphasized the joint role of explainability and generalizability, showing that integrating CNNs with XAI methods such as Grad-CAM, XGradCAM, and LayerCAM enhances visual interpretability and diagnostic consistency across multiple medical datasets.  
~\cite{unknown} presents a focused scoping review on explainability in breast cancer detection identifying SHAP as the most widely used and clinically interpretable method, particularly suited to feature-level analysis and biomarker attribution in survival and risk prediction.

Despite the steady progress of XAI in medicine, most existing frameworks emphasize saliency visualization or heuristic attribution without enforcing \emph{minimality}—the principle that an explanation should contain only the smallest subset of evidence sufficient to preserve a model’s decision.  

Our proposed work \textbf{Med-CAM}, addresses this limitation by formulating explanation generation as an activation-matching inverse problem: we seek the minimal pre-image that preserves the classifier’s internal activations and decision, thereby producing compact, faithful, and diagnostically meaningful evidence maps. 

\begin{figure*}[t]
    \centering
    \includegraphics[width=0.85\linewidth]{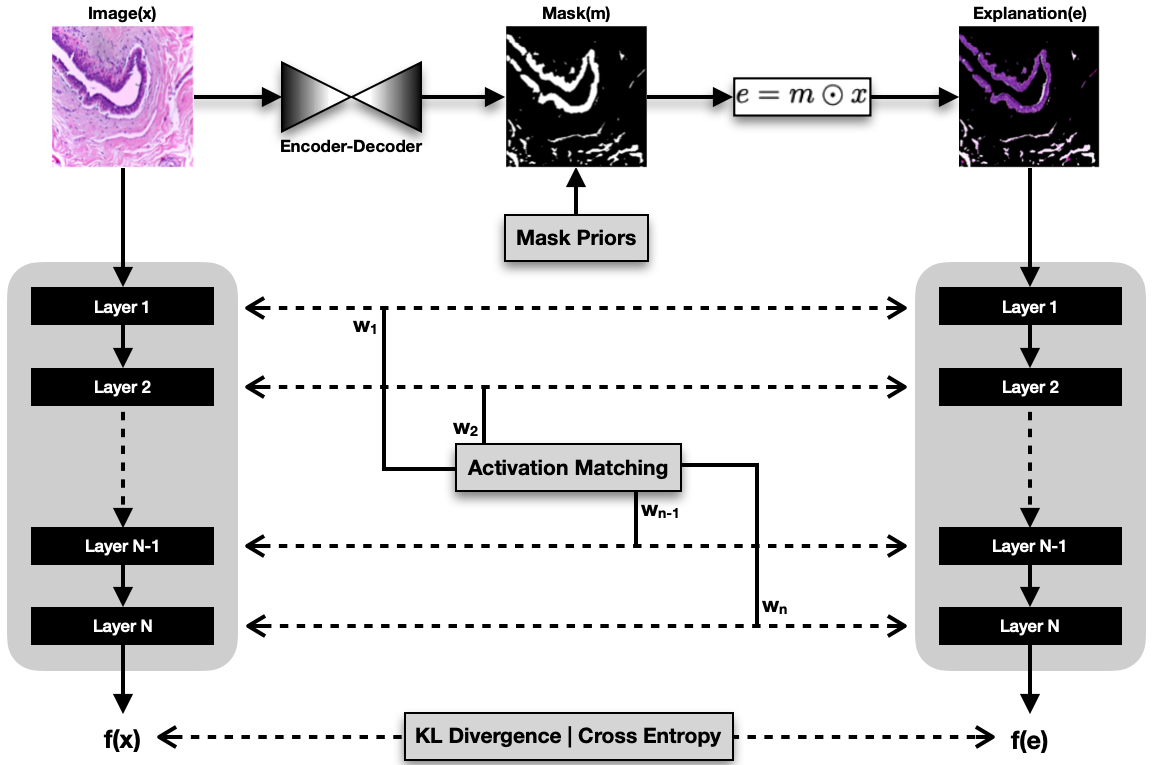}
    \caption{Overview of the Med-CAM framework. Given an input medical image \(x\), a lightweight U-Net produces a binary mask \(m\), yielding the masked explanation \(e = m \odot x\). The mask highlights the minimal diagnostic evidence required to preserve the model’s decision.}
    \label{fig:medcam_pipeline}
\end{figure*}

\section{Methodology}

Per-image explanation generation can be viewed as an \emph{inverse problem}, aiming to recover a decision-wise conclusive and critical pre-image—a minimal subset of the input—that is sufficient to sustain a model’s prediction. While the region outside this subset acts as a distractor, containing information that does not contribute meaningfully to the decision and may even obscure it.

\textbf{Med-CAM} aims to generate minimal and faithful evidence maps that explain the diagnostic decisions of a frozen medical classifier \(f\) on any given image \(x\). To achieve this, we train a lightweight U-Net that outputs a binary mask \(m\), where the masked explanation is defined as
\(
e = m \odot x ,
\)
and non-critical regions are suppressed as shown in Figure \ref{fig:medcam_pipeline}. The autoencoder is optimized individually for each image, requiring only a few seconds to converge, thus adapting dynamically to case-specific diagnostic cues.  
Training is guided by a composite loss function that integrates \emph{activation matching}, \emph{output fidelity}, and \emph{mask priors} for minimality, along with a robustness constraint ensuring clinical sufficiency and stability.

\subsection{Weighted Activation Matching}
The core principle of Med-CAM is that the explanation must preserve the classifier’s decision process at both the output and feature levels. This is enforced through activation alignment, distributional consistency and label match between the original image \(x\) and the masked explanation \(e\).

\textbf{Activation Matching Loss:} To ensure that the explanation elicits the same internal responses as the original input, we minimize
\[
\mathcal{L}_{\text{act}} = \sum_{\ell} \alpha_\ell \, d\!\big(\phi_\ell(x), \phi_\ell(e)\big),
\]
where \(\phi_\ell(\cdot)\) denotes the post-ReLU activations at layer \(\ell\), \(d\) is a distance metric (mean squared error or cosine distance), and \(\alpha_\ell\) controls per-layer weighting.  
This loss ensures that \(e\) reproduces the hierarchical activations responsible for diagnostic reasoning—such as tissue texture patterns, lesion morphology, or intensity gradients.

\textbf{KL Divergence Loss:} To preserve the classifier’s probabilistic output distribution, we minimize the Kullback–Leibler divergence between the softmax outputs of the original and masked images:
\[
\mathcal{L}_{\text{KL}} = D_{\text{KL}}\!\big( \text{softmax}(f(x)) \,\|\, \text{softmax}(f(e)) \big).
\]
This encourages the masked explanation to produce the same confidence distribution as the original image, ensuring diagnostic consistency.

\textbf{Cross-Entropy Loss:} To explicitly retain the predicted class label, we minimize

\[
\mathcal{L}_{\text{CE}} = - \log p_{f(e)}(y),
\]

where \(y\) is the top-1 class predicted by \(f(x)\).  
This enforces decision-level faithfulness: the masked explanation must independently yield the same diagnosis.
\begin{figure*}[t]
    \centering
    \includegraphics[width=1\linewidth]{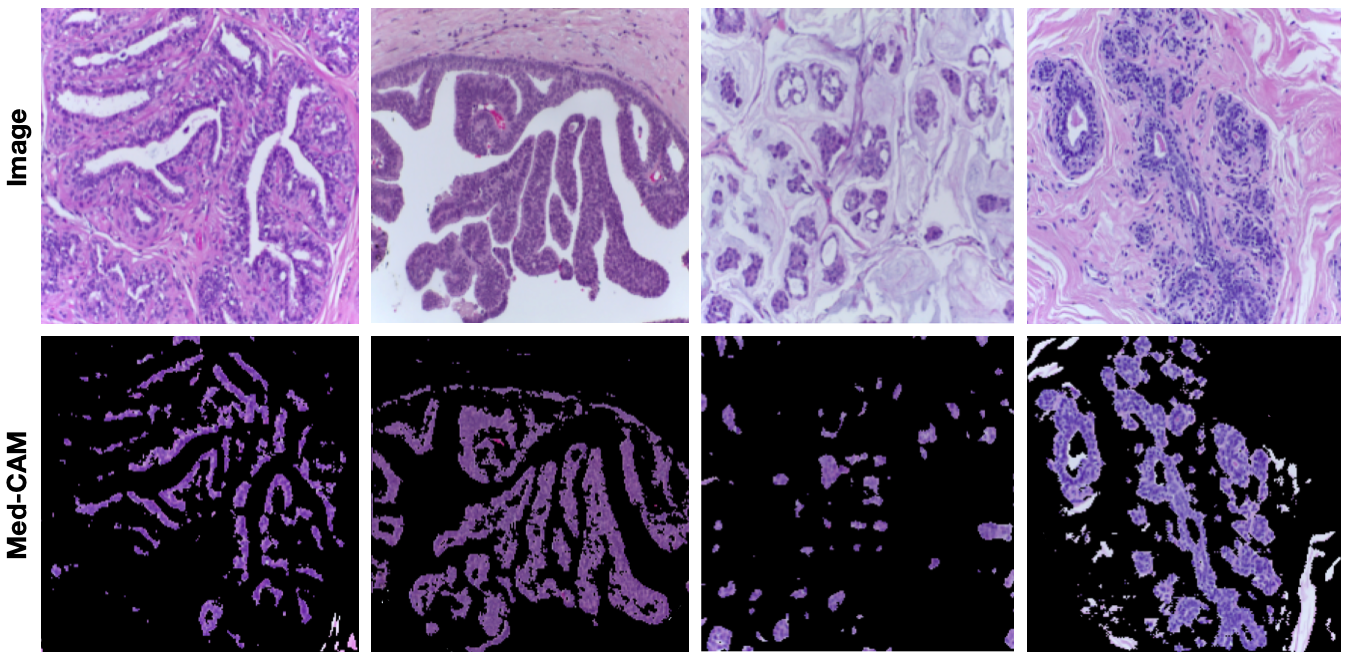}
    \caption{Med-CAM explanations on BACH H\&E breast cancer histopathology slides using a ViT-16 classifier.}
    \label{fig:bach_results}
\end{figure*}

\subsection{Mask Priors for Minimal Evidence}

To generate interpretable evidence maps that highlight diagnostically relevant regions, we impose priors on the mask \(m\).

\textbf{Area Loss for Sparsity:}
\[
\mathcal{L}_{\text{area}} = \|m\|_1.
\]
This term penalizes large active regions, encouraging Med-CAM to use the fewest possible pixels necessary for preserving the model’s decision corresponding to the principle of \emph{minimality}.

\textbf{Binarization Loss for Crispness:}
\[
\mathcal{L}_{\text{bin}} = \|m - m^2\|_1.
\]
This drives the mask toward binary values (0 or 1), producing clear, interpretable evidence maps rather than diffuse saliency heatmaps. 

\textbf{Total Variation Loss for Smoothness:}
\[
\mathcal{L}_{\text{tv}} = \sum_{i,j} \big(|m_{i,j} - m_{i+1,j}| + |m_{i,j} - m_{i,j+1}|\big).
\]
This regularizes the spatial structure of the mask, promoting contiguous evidence regions and suppressing isolated activations critical for delineating anatomical boundaries.

\subsection{Abductive Robustness Constraint}

Minimality alone does not ensure that the explanation is clinically sufficient.  
To enforce robustness, we introduce an abductive constraint ensuring that randomizing irrelevant regions does not alter the model’s decision.  
Given a perturbed background \(r\), we define \(
\tilde{e} = m \odot x + (1 - m) \odot r ,
\) and require that \(f(\tilde{e})\) yield the same label as \(f(x)\) via the loss
\[
\mathcal{L}_{\text{rob}} = - \log p_{f(\tilde{e})}(y).
\]
This ensures that explanations are faithful even when non-evidence regions are replaced with random or domain-shifted content—an important criterion for robustness in medical interpretation.

\subsection{Training Objective}

The Med-CAM U-Net is optimized using a composite loss \(\mathcal{L}_{\text{EXP}}\) that enforces activation alignment, minimality, and robustness. For clarity, we group the terms as:

\[
\begin{aligned}
\mathcal{L}_{\text{AM}} \; &=\;
\lambda_{\text{act}}\,\mathcal{L}_{\text{act}}
+\lambda_{\text{CE}}\,\mathcal{L}_{\text{CE}}
+\lambda_{\text{KL}}\,\mathcal{L}_{\text{KL}},
\\
\mathcal{L}_{\text{MIN}} \; &=\;
\lambda_{\text{area}}\,\mathcal{L}_{\text{area}}
+\lambda_{\text{bin}}\,\mathcal{L}_{\text{bin}}
+\lambda_{\text{tv}}\,\mathcal{L}_{\text{tv}},
\\
\mathcal{L}_{\text{ROB}} \; &=\;
\lambda_{\text{rob}}\,\mathcal{L}_{\text{rob}}.
\end{aligned}
\]

The complete objective is:
\[
\mathcal{L}_{\text{EXP}}
=\mathcal{L}_{\text{AM}}
+\mathcal{L}_{\text{MIN}}
+\mathcal{L}_{\text{ROB}}.
\]

With appropriate coefficient choices \(\{\lambda_{\cdot}\}\), Med-CAM learns binary, sparse, and spatially coherent masks that retain the classifier’s decision and internal activations while suppressing irrelevant context. Unlike Grad-CAM or attention maps, which provide coarse regions of relative importance, Med-CAM produces conclusive pixel-level evidence maps that isolate the minimal diagnostic cues essential to the model’s prediction.

\begin{figure*}[t]
\centering
\includegraphics[width=1\linewidth]{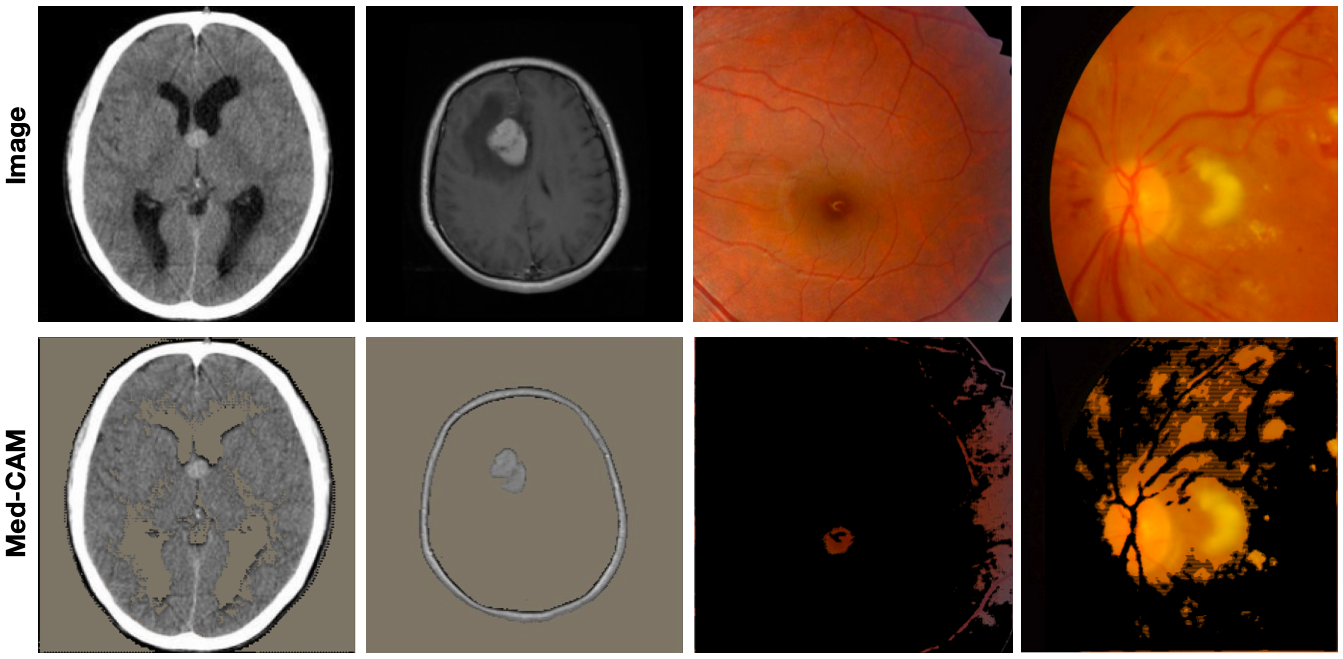}
\caption{Med-CAM explanations on brain tumor MRI (left) and IDRiD retinal fundus images (right).}
\label{fig:mri_results}
\end{figure*}

\section{Results}
Med-CAM is model-agnostic and operates on any frozen classifier and any
seen or unseen image at inference time. Because the U-Net~\cite{ronneberger2015unetconvolutionalnetworksbiomedical} explainer is trained
per-image in a few seconds, the framework naturally adapts to 
case-specific visual cues without requiring dataset-level retraining. 
We evaluate Med-CAM on four diverse medical imaging modalities—dermatology, 
histopathology, MRI, and retinal fundus—using four architectures: 
ViT-16~\cite{dosovitskiy2021imageworth16x16words} on BACH ~\cite{Aresta_2019}(89\% test accuracy), ConvNeXt-Small~\cite{liu2022convnet2020s} on HAM10000 ~\cite{Tschandl_2018}(85\%), 
ResNet-18 ~\cite{he2015deepresiduallearningimage} on IDRiD ~\cite{data3030025} (79\%), and MobileNet-V2 ~\cite{sandler2019mobilenetv2invertedresidualslinear} on Brain Tumor MRI (82\%). 

Figure~\ref{fig:bach_results} shows an image from each BACH class 
(\emph{Normal, Benign, In Situ, Invasive}) with corresponding Med-CAM explanations. Histology images exhibit highly irregular, multi-scale patterns, making explanation generation particularly challenging. Med-CAM successfully adapts to these diverse morphologies, highlighting luminal structures, glandular boundaries, and malignant epithelial clusters. Using relative weights of \(10{:}100{:}10\) for activation matching, minimality, and robustness, we observe that Med-CAM consistently isolates concise diagnostic evidence improving classifier’s confidence from an average of 85\% on the original images to 96\% on the Med-CAM explanations.

In Figure~\ref{fig:mri_results} for MRI, we include one healthy and one tumor-containing slice. Med-CAM highlights the outer brain boundary and central parenchyma for normal cases, and selectively isolates only the tumor core and surrounding edema for pathological cases. For IDRiD, we show images from retinopathy grades 2 and 4. Grade 2 exhibits mild lesions, and Med-CAM correspondingly identifies a very small, compact set of pixels capturing microaneurysms and early exudates. Grade 4 contains large, clinically significant lesions highlighted by  Med-CAM.

Figure~\ref{fig:ham_gradcam} shows results for four HAM10000 classes 
(\textit{Melanoma, Nevus, Vascular Lesion, Dermatofibroma}) in which Med-CAM produces explanations that almost resemble segmentation masks--highlighting pigment networks, lesion borders, vascular blobs, and firm nodular structures with remarkable precision.
With relative loss weights of \(10{:}150{:}20\), the evidence masks remain 
compact but highly informative, increasing classification confidence from 
an average of 75\% to 93\% after masking.  

\section{Comparisons}
\begin{figure*}[h]
\centering
\includegraphics[width=\linewidth]{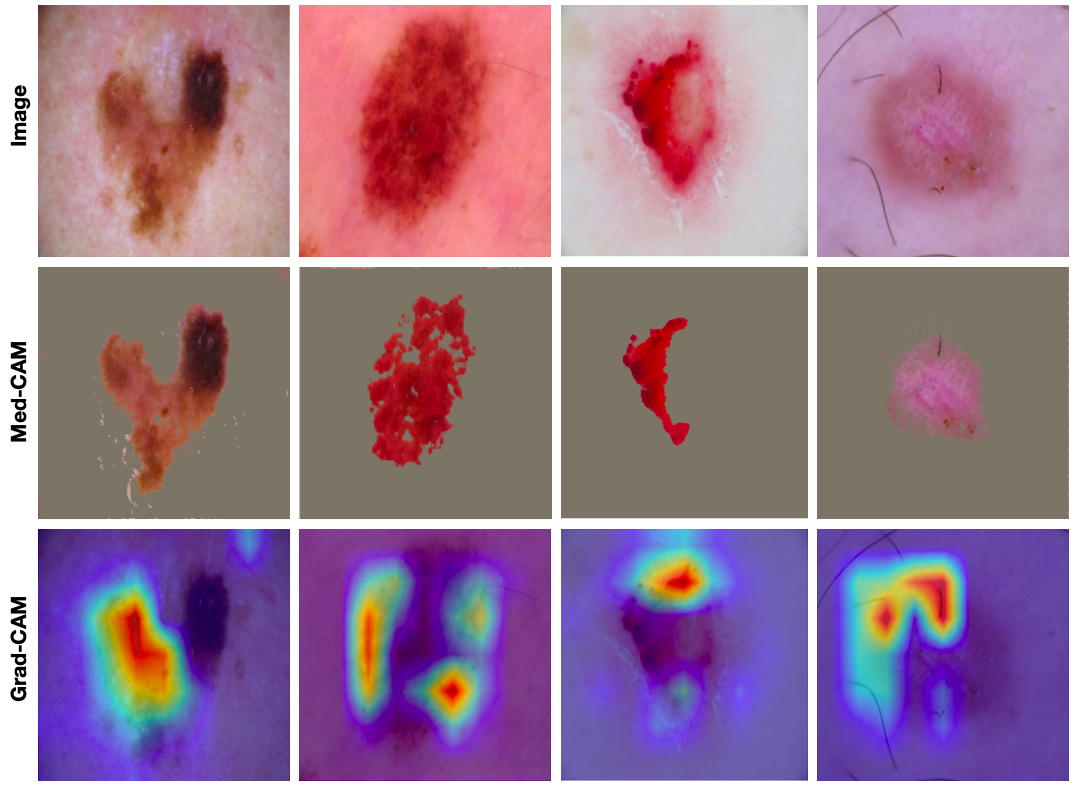}
\caption{\textbf{Med-CAM vs Grad-CAM on HAM10000.} 
}
\label{fig:ham_gradcam}
\end{figure*}

We compare Med-CAM against the widely used Grad-CAM on the 
HAM10000 dataset (Figure~\ref{fig:ham_gradcam}). Grad-CAM heatmaps highlight large, smooth regions of relative importance but lack spatial awareness of fine lesion structure. Grad-CAM explanations fragment into multiple disconnected blobs despite the lesion being compact. Their activations frequently diffuse into background skin, illumination artifacts, vellus hair, and regions not diagnostic of the underlying condition. Moreover, 

Med-CAM, in contrast, produces crisp, binary explanations with sharply defined 
boundaries that adhere to the exact lesion morphology. The method precisely 
captures irregular lesion outlines, pigment asymmetry, and subtle textural cues 
essential for melanoma and nevus differentiation. Unlike Grad-CAM, Med-CAM’s 
evidence masks are minimal yet guaranteed to preserve the classifier's original 
diagnosis. Empirically, Grad-CAM often spreads its attention across both lesion and non-lesion regions, while Med-CAM isolates only the discriminative pixels, returning a compact pre-image that is decision-equivalent to the full input.

\section{Conclusion}
\label{sec:conclusion}
We introduced Med-CAM, a framework to generate minimal evidential explanations for medical image classifiers. Med-CAM consistently isolates the smallest diagnostically sufficient regions while preserving the model’s decision. By offering precise, compact, and clinically  verifiable evidence maps, Med-CAM advances the reliability and interpretability of medical AI systems. 
{
    \small
    \bibliographystyle{ieeenat_fullname}
    \bibliography{main}
}


\end{document}